\documentclass{article}
\usepackage[preprint,nonatbib]{Style/neurips_2020}
\usepackage{wrapfig}

\usepackage[utf8]{inputenc} 
\usepackage[T1]{fontenc}    
\usepackage{xcolor}
\usepackage{xspace}
\usepackage{amsmath}
\usepackage{float}
\usepackage{dsfont}
\usepackage{subcaption}
\usepackage{pict2e}
\usepackage{booktabs}
\usepackage[pdftex]{graphicx}
\usepackage{Macros/slashbox}

\newcommand{\comment}[1]{}

\newcommand{\semspace}{\ensuremath{\Theta}}
\newcommand{\seminp}{\ensuremath{\theta}}
\newcommand{\semce}{\ensuremath{\tilde{\seminp}}}
\newcommand{\semmod}{\ensuremath{\pi}}
\newcommand{\inp}{\ensuremath{x}}
\newcommand{\outp}{\ensuremath{y}}
\newcommand{\render}{\ensuremath{R}}

\newcounter{myctr}
\newenvironment{mylist}{\begin{list}{\arabic{myctr}.}
{\usecounter{myctr}
\setlength{\topsep}{1mm}\setlength{\itemsep}{0.25mm}
\setlength{\parsep}{0.1mm}
\setlength{\itemindent}{0mm}\setlength{\partopsep}{0mm}
\setlength{\labelwidth}{15mm}
\setlength{\leftmargin}{4mm}}}{\end{list}}

\usepackage{amssymb}
\usepackage{amsmath}
\usepackage{amsthm}
\usepackage{balance}
\usepackage{color}
\usepackage{enumerate}
\usepackage{enumitem}
\usepackage{epstopdf}
\usepackage{etoolbox}
\usepackage{hyperref}
\usepackage{multirow}

\hypersetup{
    colorlinks,
    linkcolor={red},
    citecolor={black},
    urlcolor={black}
}

\newcommand{\Real}{{\mathbb R}}

\DeclareMathOperator{\bftheta}{{\bf \theta}}

\DeclareMathOperator*{\signf}{sign} 
\DeclareMathOperator*{\argmax}{arg\,max}

\newtheorem*{theorem*}{Theorem}
\newtheorem*{definition*}{Definition}

\newcommand*{\ie}{{\em i.e.,}\@\xspace}
\newcommand*{\etal}{{\em et al.}\@\xspace}
\newcommand*{\eg}{{\em e.g.,}\@\xspace}
\newcommand*{\cars}{{\em color}\@\xspace}
\newcommand*{\sky}{{\em weather}\@\xspace}
\newcommand*{\foliage}{{\em foliage}\@\xspace}
\newcommand*{\translate}{{\em translate}\@\xspace}
\newcommand*{\rotate}{{\em rotate}\@\xspace}
\newcommand*{\FFD}{{\em mesh}\@\xspace}
\newcommand*{\TEMPLATE}{{\tt TEMPLATE}\@\xspace}
\newcommand*{\SAE}{{semantic counterexample}\@\xspace}
\newcommand*{\SAEs}{{semantic counterexamples}\@\xspace}

\newcommand*{\pose}{{\em pose}\@\xspace}
\newcommand*{\vertex}{{\em vertex}\@\xspace}
\newcommand*{\lighting}{{\em lighting}\@\xspace}

\newcommand{\varun}[1]{\textcolor{blue}{Varun: #1}}
\newcommand\UJ[1]{\textcolor{red}{UJ:#1}}

\newcommand\somesh[1]{\textcolor{purple}{Somesh:#1}}

\newcommand\sanjit[1]{\textcolor{teal}{Sanjit:#1}}

\title{Analyzing and Improving Neural Networks by Generating Semantic Counterexamples through Differentiable Rendering}
\author{Lakshya Jain, Varun Chandrasekaran$^*$, Uyeong Jang$^*$, Wilson Wu, Andrew Lee, \\  \bf{Andy Yan, Steven Chen, Somesh Jha$^*$, Sanjit A. Seshia}\\University of California, Berkeley, $^*$University of Wisconsin-Madison}

\begin{document}
\maketitle
\begin{abstract}

Even as deep neural networks (DNNs) have achieved remarkable success on vision-related tasks, their performance is brittle to transformations in the input. Of particular interest are
semantic transformations that model changes that have a basis in the physical
world, such as rotations, translations, changes in lighting or camera pose. In this paper, we show how {\em differentiable rendering} can be utilized to
generate images that are {\em informative}, yet {\em realistic}, and which can be used to analyze DNN performance and improve its robustness through data augmentation. Given a differentiable renderer and a DNN, we show how to use off-the-shelf attacks from adversarial machine learning to generate {\em semantic counterexamples} --- images where semantic features are changed as to produce misclassifications or misdetections.  
We validate our approach on DNNs for image classification and object detection. 
For classification, we show that semantic counterexamples, when used to augment the dataset, (i) improve generalization performance (ii) enhance robustness to semantic transformations, and (iii) transfer between models. Additionally, in comparison to sampling-based semantic augmentation, our technique generates more informative data in a sample efficient manner.

\end{abstract}

\section{Introduction}
\label{sec:intro}

Machine Learning (ML) models, such as deep neural networks (DNNs), have shown remarkable success in several domains, including visual tasks such as image classification and object detection.
Thus, ML models have started being used in safety-critical applications
such as in autonomous driving and other cyber-physical systems (CPS).
At the same time, it has been well documented that DNN performance can be
brittle to small perturbations of the input data~\cite{engstrom2019exploring,athalye2017synthesizing,papernot2016limitations,FGSM,carlini2017towards,Madry,eykholt2018robust}.
Such brittleness of ML models in safety-critical CPS can have disastrous consequences.

Semantic modifications to input images, capturing changes that have a basis
in the physical world, are particularly important in CPS.
These modifications ({\em c.f.} Appendix~\ref{app:samples}) include translations, rotations, changes in lighting,
contrast, or color, changes in camera pose, time of day, object deformations. They capture perturbations to the environment of the ML-based system that
are {\em semantically meaningful}, and more likely to occur in reality. 
Small semantic modifications to the input should not negatively affect the output of the ML model; in other words, we want the ML model to possess {\em semantic robustness}. For example, for an object detector that must identify cars in an image and draw bounding boxes around them, the output should remain unchanged to a change of car colors. Inputs that violate semantic robustness are termed as {\em \SAEs}. In spite of the impressive depth and volume of work on adversarial ML applied to computer vision, most of that literature focuses on pixel-level transformations to the input~\cite{papernot2016limitations,FGSM,carlini2017towards,Madry,moosavi2016deepfool}. There is a need for effective techniques for generating semantic
counterexamples and for using them to improve the semantic robustness of the ML model (DNN).

\begin{figure}[ht]
\begin{minipage}[b]{0.5\textwidth}
\includegraphics[width=\textwidth]{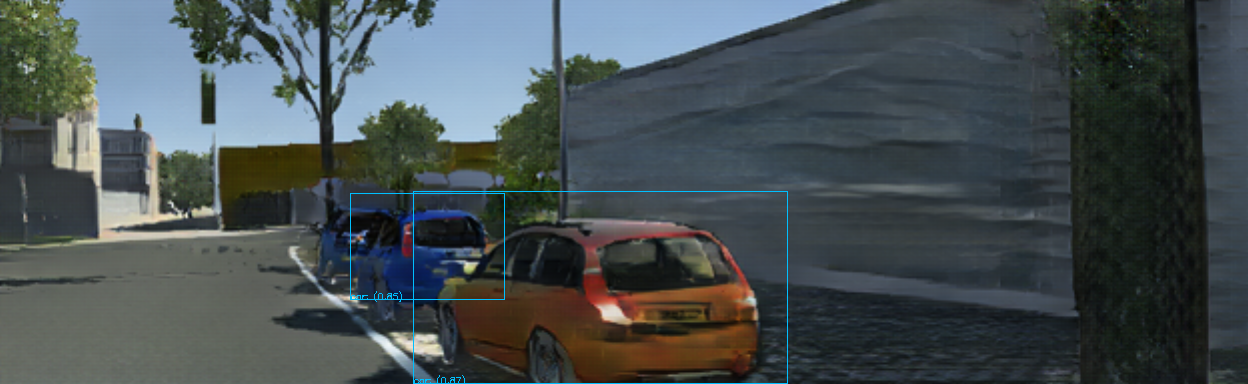}
\end{minipage}
\begin{minipage}[b]{0.5\textwidth}
\includegraphics[width=\textwidth]{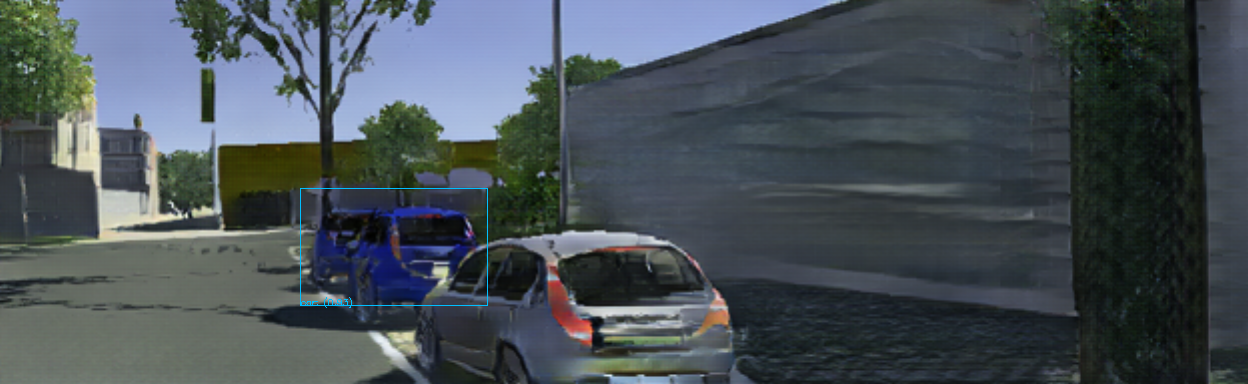}
\end{minipage}
\caption{\small \textbf{Semantic counterexamples for object detection.} Benign re-rendered {VKITTI} image (left), \SAE generated by our approach (right). The semantic changes involve changes in car positions, orientations, and color. This causes the network to fail to detect the grey car in the immediate foreground.}
\vspace{-6mm}
\label{fig:obj-detection-sae}
\end{figure}

In this paper, we address this need through an approach that leverages advances in adversarial ML and differentiable rendering.
We begin by defining a semantic feature space capturing features of the environment that, together with a rendering process, determine the input image. 
In semantic robustness, we are interested in exploring modifications to points in the semantic feature space that produce incorrect outputs.
Adversarial ML techniques currently provide an effective way to produce such
modifications in the pixel space for images. We show how advances in differentiable rendering allow us to use off-the-shelf attacks from adversarial ML literature to generate \SAEs in a sample-efficient fashion.
For object detection, we show how our approach can produce semantically-meaningful images that are misdetected (see example in Figure~\ref{fig:obj-detection-sae}). 
For classification, we show how dataset augmentation with \SAEs results in 
(i) improved generalization performance, 
and
(ii) enhanced semantic robustness across adversarial attacks.
We also observe the transferability of \SAEs between classification models.

We validate our approach on DNNs for both image classification and object detection. Our results show that, in comparison to sampling-based semantic augmentation, our technique generates more informative data given a query budget. Our empirical evaluation shows how our approach can be implemented using three candidate attack algorithms (iterative FGSM~\cite{kurakin2016adversarial}, a variant of PGD~\cite{Madry}, and Carlini-Wagner~\cite{carlini2017towards}) and two differentiable rendering frameworks (3D-SDN~\cite{3dsdn2018} and Redner~\cite{redner}) utilizing two popular synthetic datasets (VKITTI~\cite{vkitti} and ShapeNet~\cite{chang2015shapenet}) and analyzing two different object detectors (SqueezeDet~\cite{squeezedet} and YOLOv3~\cite{redmon2018yolov3}) and two image classification networks (ResNet-50~\cite{resnet} and VGG-16~\cite{vgg}). Our results suggest that, for image classification, \SAEs can induce upto 50 percentage point accuracy degradation whilst maintaining realism ({\em c.f.} \S~\ref{realism}). These counterexamples are highly informative ({\em c.f.} \S~\ref{gain}), and can be used to improve semantic robustness against those generated using different strategies ({\em c.f.} \S~\ref{robustness}).

{\bf Related Work.}
Our work builds upon the literature on differentiable rendering and adversarial ML, which is reviewed in depth in Appendix~\ref{app:related}. In particular, we use Redner~\cite{li2018differentiable}, 
a general-purpose differentiable ray tracer for our image classification experiments,
and 3D-SDN~\cite{3dsdn2018} for our object detection experiments.
The extensive prior work on Adversarial ML (\eg~\cite{Madry,FGSM,carlini2017towards,moosavi2016deepfool,papernot2016limitations}) 
focuses on generating norm-bounded pixel-level changes to input images; however, these manipulations are not usually realizable in the real world. Generating real world adversarial examples has resulted in several efforts focused on particular semantic features,
including physical world attacks~\cite{eykholt2018robust},
changing color~\cite{bhattad2020unrestricted} and texture~\cite{Texture,bhattad2020unrestricted},
modifying the spatial orientation of images~\cite{engstrom2019exploring}, and
changing pixel intensities~\cite{sharif2016accessorize}. 
Song \etal \cite{songGAN} utilize GANs to construct examples from scratch instead of taking existing datapoints and perturbing them, and demonstrate some semantically meaningful changes. 
Approaches that involve explicitly sampling the semantic parameter space are slow, imprecise, and expensive~\cite{SAESeshia,dreossi-ijcai18}. 
The most closely related work is by Xiao \etal~\cite{Xiao_2019_CVPR} and Qiu \etal~\cite{SemanticAdv}; the former uses a differentiable renderer to induce changes in shapes and textures, while the latter induces a different set of semantic changes using a generative model. Our approach stands out by being general, in that it can apply to any semantic features as long as we can render from those features using a differentiable renderer. Further, we show how our generated \SAEs can be effectively used for data augmentation and that they transfer between DNNs (used for classification).

{\bf Roadmap.} Our approach for generating \SAEs is described in \S~\ref{sec:sae} (due to space constraints, some details are left to Appendix~\ref{constructions}). \S~\ref{sec:quality} provides a metric for measuring the {\em informativeness} of generated examples. Implementation details and evaluation appear in \S~\ref{sec:eval}.

\section{Semantic Counterexamples}
\label{sec:sae}

\begin{wrapfigure}{r}{3.2in}
\centering
\includegraphics[width=3in]{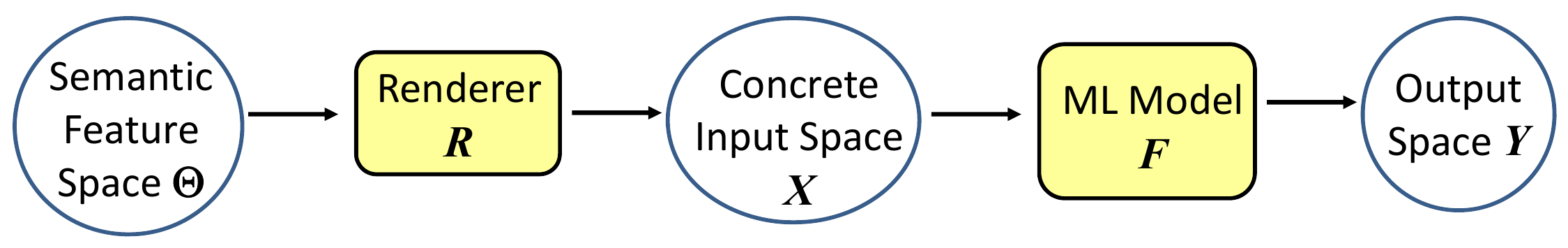}
\caption{\small \label{fig:overview}Our approach analyzes the composition of a differentiable renderer $R$ and an ML model $F$.}
\end{wrapfigure}

The components of the problem considered in this paper are shown in Figure~\ref{fig:overview}. 
We begin by defining a semantic feature space $\semspace$ capturing the features of the environment  that determine the input image.
An element of $\semspace$ is a vector of semantic features describing the environment
of the ML model, including the pose of objects, their color, texture, and other characteristics,
camera pose, lighting, time of day and weather conditions, background of the scene, etc.
Given a point $\seminp \in \semspace$, a process $\render$ (modeling rendering, simulation, or 
camera capture) produces an input image $\inp$. 
An ML model (such as a DNN) $F$ then maps $\inp$ to an output $\outp$.
In this section, we define the notions of semantic robustness and semantic counterexamples in this context. We then show how
a combination of differentiable rendering and off-the-shelf adversarial-ML techniques can be used to find semantic
counterexamples. We begin by defining some basic notation.

\subsection{Notation}
\label{notation}

Consider a space $Z$ of the form $X \times Y$, where $X$ is the sample space and $Y$ is the set of labels. 
Each point $x \in X$ is a {\em concrete input vector}, \eg a pixel-level encoding of an image; we will assume 
that $X = \Real^n$. 
Let $H$ be a hypothesis space (\eg weights of a DNN). 
We assume a loss function $\ell: H \times Z \mapsto \Real$ such that that given a hypothesis $w \in H$ 
and a labeled data point $(x,y) \in Z$, the loss is $\ell_w(x,y)$.  
The machine learning model is a function $F$ from $\Real^n$ to $Y$; in this paper, we focus on ML models that
are DNNs performing object detection or object classification tasks.
Sometimes, to emphasize that the ML model depends on a hypothesis $w \in H$,  
we will denote it as $F_w$ (if $w$ is clear from the context, we will simply write $F$, and use $\ell_F(x,y)$ in place of $\ell_w(x,y)$). 
When $F$ is a classifier, the {\em softmax} output corresponding to $F$ is denoted by $s(F)$ and 
is a function $s(F): X \rightarrow \Delta (Y)$, where $\Delta(Y)$ is the set of distributions over $Y$. 
Typically, $F(x)$ is defined as $\argmax_{y \in Y} s(F)(y)$. The notation for object detection is similar, and is described in Appendix~\ref{app:3dsdn_notation}.


\subsection{Semantic Robustness}
\label{sec:sem-robustness}
The semantic robustness property captures the requirement that a small change in the semantic feature space
must only result in a small change in the output space. Given $\seminp \in \semspace$,
a renderer $\render$ and an ML model $F$, we formulate it abstractly as follows:
$$
\forall \seminp' \in \semspace, \, \forall \inp, \inp' \in X  \, . \; 
[ \seminp \approx_{\semspace} \seminp' \land R(\seminp) = \inp \land R(\seminp') = \inp' ]
\Rightarrow
[ F(\inp) \approx F(\inp') ]
$$
where $\approx_\semspace$ specifies that $\seminp$ and $\seminp'$ are {\em close}
in the semantic space, while $\approx$ specifies that the outputs
are close/unchanged as required by the application.

In practice, the relation $\approx_\semspace$ is typically defined as having $\seminp'$ within an $\varepsilon$
neighborhood of $\seminp$ using a suitable norm $\mu$ such as as $l_\infty$, $l_0$, $l_1$, or $l_p$ ($p \geq 2$).
Similarly $\approx$ is defined as either requiring $F(\inp) = F(\inp')$ or having them within a specified
neighborhood of each other.

\subsection{Traditional Adversarial Examples}
\label{trad}

A common goal in adversarial ML is for an adversary $A$ to take any input vector 
$x \in \Real^n$ and produce a minimally altered version denoted by $A(x)$, that has the property of being misclassified by a classifier $F$. 
Usually this is formulated as the following optimization problem:
\[
\begin{array}{lr}
\min_{\delta \in \Real^n}  & \mu ( \delta ) \\
\mbox{such that} & F(x+\delta) \not= F(x) 
\end{array}
\]
where $\mu$ is a norm on $\Real^n$, such as $l_\infty$, $l_0$, $l_1$, or $l_p$ ($p \geq 2$). If $\delta$ is the solution of the optimization problem given above, then the adversarial example $A(x) \; = \; x+\delta$.

Common examples of adversarial attacks in the literature include the
{\em fast gradient sign method} (FGSM)~\cite{FGSM},
{\em projected gradient descent} (PGD)~\cite{Madry}, and
the {\em Carlini-Wagner method} (CW)~\cite{carlini2017towards}.

\subsection{Semantic Counterexamples}
\label{map}

Given a semantic feature vector $\seminp \in \semspace$, 
a {\em semantic counterexample} $\semce$ is an element of $\semspace$ in an $\varepsilon$
neighborhood of $\seminp$ that violates the semantic robustness property. 
For example, if $\seminp$ includes the position of a car, its color, and the camera pose,
$\semce$ can be a small change to the camera pose that causes $F$ to misidentify the car. 

Some differentiable renderers, such as Redner~\cite{redner}, directly generate an image $x$
from a semantic feature vector $\seminp$. However, others, such as 3D-SDN~\cite{3dsdn2018}
first de-render an input image $x$ to $\seminp$ and then re-render.
In either case, the key aspects are that the perturbation is to be performed on $\seminp$ and the
function $\render$ is differentiable.

\comment{
We assume that there is a special identity element in $\semspace$ (which we call $\perp$) such that $\tau(x,\perp) = x$. 

As discussed above, we are also given a norm $\mu$ over $\semspace$.
We assume that $\tau(x,\theta)$ is differentiable in $\theta$.
We also assume that there is an operator $\oplus$ called the aggregation 
operator (similar to addition in $\Real^n$), but in the parameter
space $\Theta$. The precise axioms satisfied by $\oplus$ depends
on $\Theta$, but one axiom we require in our construction is the following:
\begin{eqnarray*}
\tau(x, \bftheta_1 \oplus \bftheta_2) & = & 
\tau(\tau(x,\bftheta_1),\bftheta_2)
\end{eqnarray*}
} 


Given this, we formulate the procedure of finding a semantic counterexample also as
an optimization problem, as follows:
\[
\begin{array}{lr}
\min_{\semmod \in \semspace}  & \nu ( \semmod ) \\
\mbox{such that} & F(R(\seminp + \semmod)) \not= F(R(\seminp)) 
\end{array}
\]
where $\nu$ is a norm on $\semspace$. 
In other words, we want to find a {\em small perturbation} in the semantic space $\semspace$ 
that will misclassify the sample.
If the result of the optimization is $\semmod^*$, then the generated semantic counterexample is
$\semce = \seminp + \semmod^*$.\footnote{We note that the semantic parameter space is not homogeneous and it is unclear if one function can be used to capture a suitable notion of distance. Not only should the norm measure the changes in the semantic space, it should also approximate human perception. Additionally, in practice the ``+'' operator can be more complex for semantic parameters such as weather or background.} 

Let $A(F,x)$ be an attack algorithm to generate traditional adversarial examples (as defined in \S~\ref{trad}). We 
outline a general technique for transforming $A$ to the semantic equivalent $sA$ to generate semantic adversarial examples (henceforth referred to as \SAEs). The technique
works as by transforming algorithm $A$ using the following rules:
\begin{mylist}
\item Replace $\delta$ with $\semmod$.
\item Replace $x+\delta$ with $\seminp+\semmod$.
\item Use chain rule to compute the gradients of terms, \eg the loss function, that involve $\render(\seminp)$.
\end{mylist}

We detail some of the specific constructions we use in our experiments in Appendix~\ref{constructions}.

\section{Augmentation Quality}
\label{sec:quality}

Various strategies, such as random sampling in the semantic parameter space, can be utilized to generate  samples
for augmentation. Suppose there is a dataset $D$ and we have two examples for augmentation $(x,y)$ and 
$(x',y')$. Which one is better for augmentation?
In this section, we formalize the notion of {\em informativeness} to address
this question. Different variants of such metrics have been extensively studied~\cite{ghorbani2020distributional}. We highlight the intuition behind formalizing our metric in Appendix~\ref{notes_augmentation}.

\subsection{Entropy \& Information Worth}
\label{entropy}

A common measurement that is used to measure the {\em goodness} of a classifier's predictive capabilities makes use of the concept of Shannon entropy~\cite{shannon, quinlan} which measures the unpredictability of outcome from a given set.
Let the label set $Y$ have $L$ labels $\{l_1,\cdots, l_L \}$.
Given set of datapoints $D=\{(x_1, y_1),\cdots, (x_m, y_m)\}$ where $y_t\in Y$ is the true label of a point $x_t$, a classifier $F$ on the set $D$ divides the set into subsets $D_1, \ldots, D_L$ where $D_i$ contains points whose predicted label is $l_i$. Let $E_i$ be the entropy measured on each subset $D_i$ of size $m_i$ \ie $E_i = -\sum_{j\in [L]} p_j^i\log p_j^i$, where $p_j^i$ is the estimated probability of observing the class $j$ in the subset $D_i$.

If the entropy $E_i$ for each subset $D_i$ is low, then the classifier divides $D$ in a well-organized manner. Thus, the following weighted average of $E_i$'s (henceforth referred to as {\em information worth}) is used to estimated the quality of a classifier \ie  $\hat E = \sum_{i\in[L]}\gamma_i \cdot E_i$, where $\gamma_i = \frac{|D_i|}{|D|} = \frac{m_i}{m}$. 


We also want to make use $\hat E$ as a measure of the quality of augmented data. As long as the datapoints used for augmentation are realistic, it is better to have datapoints that do not agree with the current classifier, as we expect the re-training to improve the classifier by fixing the incorrect classification. However, the previous definition of $\hat E$ depends only on the predictions of the classifier. Thus, it is hard to distinguish between augmenting samples generated using the methodology defined in \S~\ref{map} from those generated using random sampling, for example. 

\subsection{Incorporating Classifier Confidence \& Realism} 

In the previous definition, each point $x$ could belong to only 1 subset (\ie binary membership). However, with most DNN-based models, a point could belong to many classes (and consequently subsets) depending on its softmax values (\ie fractional membership). Thus, let $\mu_i(x)$ denote the membership of a point to subset $i$. The most natural choice of $\mu_i(x) = s(F)(x)_i$ which denotes the softmax value of $x$ for class $i$. We also assume that there exists a quantifiable measure of realism $\rho$ such that $\rho(x)\in [0,1]$ (larger value means the image is more realistic). In this work, since we generate a new datapoint $x'$ from a reference datapoint $x^*$, we first define the similarity $\sigma$ of $x'$ to $x^*$ as $\sigma(x'; x^*) = 1 - \frac{d(x', x)}{d_\text{max}}$, where $d$ is the LPIPS distance~\cite{lpips} and $d_\text{max}$ is the maximum LPIPS score achievable. This metric assigns $\rho(x^*;x^*)=1$, because when there is no perturbation, \ie $x'=x^*$, then $d(x^*,x^*) = 0$. Therefore, we define $\rho(x)$ for $x\in S$ as 
$\sigma (x; x^*)$. Since $\rho$ captures {\em realism}, if the value $\rho(x)$ is small for some data point $x$, we do not trust the classification result $F(x)$. Based on this intuition, we control the participation of each datapoint $x$ in the information worth calculation by weighting it by its realism $\rho(x)$.  These ideas are captured by using the following $p_j^i$ and $\gamma_i$ in $\hat{E}$.
\begin{align*}
    p_j^i &= \frac{\sum_{t=1}^m \rho(x_t)\mu_i(x_t)\mathds{1}[y_t=j]}{\sum_{l\in[L]}\sum_{t=1}^m \rho(x_t)\mu_i(x_t)\mathds{1}[y_t=l]} = \frac{\sum_{t=1}^m \rho(x_t)\mu_i(x_t)\mathds{1}[y_t=j]}{\sum_{t=1}^m \rho(x_t)\mu_i(x_t)}\\
    \gamma_i &= \frac{\sum_{t=1}^m \rho(x_t)\mu_i(x_t)}{\sum_{l\in[L]}\sum_{t=1}^m \rho(x_t)\mu_i(x_t)}
\end{align*}

\section{Evaluation}
\label{sec:eval}

Our evaluation is centered around answering the following questions:
\begin{mylist}
\itemsep0em
\item Are the \SAEs informative?
\item Are the \SAEs realistic? 
\item Can the \SAEs be used in a sample-efficient manner to improve model robustness?
\end{mylist}

Our experiments were performed on two servers. The first has an NVIDIA Titan GP102 GPU, 8 CPU cores, and 15GB memory. The second has 264 GB memory, 8 NVIDIA's GeForce RTX 2080 GPUs, and 48 CPU cores. Due to space constraints, we restrict our discussion to the experiments involving image classification (using VGG-16 \cite{vgg16} as the target\footnote{Experiments with ResNet-50 follow a similar trend and are detailed in Appendix~\ref{expts_resnet}}). The experiments involving object detection are detailed in Appendix~\ref{3dsdn_expts}. We observe that:

\begin{mylist}
\item Semantic counterexamples are shown to be both realistic and informative ({\em c.f.} \S~\ref{gain} and \S~\ref{realism}) using both quantitative and qualitative measures.  
\item Classification models augmented using \SAEs do not suffer from generalization degradation, but show improved robustness ({\em c.f.} \S~\ref{robustness}) across \textit{all} tested generation methods.
\item 
The \SAEs we generate are sample-efficient and transferable across networks -- \ie those generated using VGG-16 transfer to ResNet-50 ({\em c.f.} \S~\ref{transferability}).
\end{mylist}

\subsection{Implementation}
\label{sec:implementation}


For our experiments, we utilize two differentiable graphics frameworks: (i) 3D-SDN proposed by Yao \etal~\cite{3dsdn2018}, and (ii) Redner proposed by Li \etal~\cite{li2018differentiable}. Using these frameworks, we augment two popular synthetic datasets: (i) VKITTI~\cite{vkitti}: scenes from photo-realistic proxy virtual worlds used for multi-object tracking, and (ii) ShapeNet~\cite{chang2015shapenet}: 
a large-scale repository of shapes represented by 3D CAD models of objects. The former dataset is used for object detection with SqueezeDet~\cite{squeezedet} as the target, and the latter for image classification using VGG-16~\cite{vgg} and ResNet-50~\cite{resnet} as targets.

We generate \SAEs by modifying elements from 3 popular methods -- i-FGSM~\cite{kurakin2016adversarial} (iterative FGSM), a variant of PGD~\cite{Madry} (henceforth called GD), and CW~\cite{carlini2017towards}  -- to generate their semantic counterparts (with the prefix {\em s}). Our implementation with 3D-SDN involves 1000 lines of code; this involves making changes to the original code to enable end-to-end differentiation, and replacing several image manipulations with their differentiable counterparts. Our implementation with Redner~\cite{li2018differentiable} involves 618 lines of code. The code is available at \url{https://github.com/BerkeleyLearnVerify/rednercounterexamplegenerator}\footnote{We cannot open source the code using 3D-SDN as the license for the 3D-SDN code does not permit it.}. The overall pipeline is presented in Figure~\ref{fig:redner_pipeline}. We compare these \SAEs with augmenting samples generated using (i) random sampling and (ii) Halton sampling~\cite{halton} as baselines. All \SAEs generated involve modifying multiple semantic parameters ({\em c.f.} \S~\ref{combinations} in the Appendix). More details, such as the exact hyperparameters used for our generation process, the semantic parameters modified, the dataset sizes, model parameters etc. are presented in Appendix~\ref{specifics_implementation}.


\subsection{Informativeness}
\label{gain}

To understand if the \SAEs generated are useful for data augmentation, we measure the accuracy degradation they induce on the VGG-16 model they were generated from. For each baseline approach, we provide 2 ranges to sample rotation (the dominant semantic transformation, keeping the vertex translation parameter fixed). These are (i) the large range \ie [-0.75, +0.75] radians, and (ii) the small range \ie [-0.3, +0.3] radians. Each cell in Table~\ref{tab:strategies_test} contains per-class normalized accuracy, and the weighted average is in the last column. Observe that \SAEs generated using Halton sampling (in the large range) cause the most accuracy degradation (the difference between the overall accuracy in the benign setting and the overall accuracy for this particular approach). This is followed by \SAEs (sCW and si-FGSM). Note that the accuracy degradation is not uniform across all classes. For example, across all 3 methods we use, the degradation for class 0 (\text{airplane}) is much lower than for class 3 (\text{bus}). The same experiment is repeated with the ResNet-50 architecture~\cite{resnet} and reported in Appendix~\ref{expts_resnet}, Table~\ref{tab:strategies_test_resnet}. The results are slightly different, where si-FGSM induces the most accuracy degradation; this suggests that the hyperparameter choices impact the efficacy of the method (\ie si-FGSM is effective for ResNet-50 but sCW is effective for VGG-16).

\begin{table}[h]
\centering
\resizebox{\textwidth}{!}{
\begin{tabular}{c  c c c c c c c c c c c c c}
\toprule 
\multicolumn{14}{c}{\bf Class} \\
{\bf Strategy} &   0   &  1   &  2 & 3  & 4    & 5 & 6     &7 & 8 & 9 & 10  & 11   & \textbf{overall}  \\
\midrule
\midrule
\text{benign}  &  0.999 & 0.996 & 0.992 & 0.969 & 0.997 & 0.983 & 1 & 0.986 & 0.939 & 0.84 & 0.849 & 0.956 & \textbf{0.986} \\
\text{sCW} & 0.883 & 0.407 & 0.319 & 0 & 0.353 & 0.783 & 0.133 & 0.671 & 0.041 & 0.12 & 0.2 & 0.635 & \textbf{0.485}   \\
\text{si-FGSM} & 0.836 & 0.642 & 0.460 & 0 & 0.402 & 0.783 & 0.40 & 0.507 & 0 & 0.240 & 0.273 & 0.619 & \textbf{0.529} \\
\text{sGD}  &  0.917 & 0.707 & 0.589 & 0 & 0.589 & 0.883 & 0.467 & 0.767 & 0.041 & 0.160 & 0.232 & 0.781 & \textbf{0.657}   \\
\text{Random (large)} & 0.876 & 0.612 & 0.462 & 0 & 0.436 & 0.933 & 0.2 & 0.712 & 0.143 & 0.36 & 0.576 & 0.819 & {\bf 0.582}\\
\text{Random (small)} & 0.969 & 0.935	& 0.789	& 0.040	& 0.810	& 0.966	& 0.466	& 0.904	& 0.346	& 0.52	& 0.535	& 0.936	& {\bf 0.830} \\
\text{Halton (large)} & 0.744	& 0.293	& 0.142	& 0.008	& 0.017	& 0.833	& 0	& 0.657	& 0	& 0	& 0.010	& 0.476	& {\bf 0.274}\\
\text{Halton (small)} & 0.979	& 0.935	& 0.832	& 0.165	& 0.690	& 0.967	& 0.667	& 0.945	& 0.245	& 0.32	& 0.131	& 0.721	& {\bf 0.757}\\
\bottomrule
\end{tabular}
}
\caption{\small {\bf Accuracy degradation} induced by different \SAE-generation strategies. Observe that \SAEs are effective at inducing accuracy degradation (lower the {\bf overall} value, the better). All experiments are carried out using VGG-16 as the target model.}
\label{tab:strategies_test}
\end{table}
\vspace{-2mm}

When the datasets are augmented using \SAEs, the accuracy of the newly learned models are presented in Table~\ref{tab:accuracy_improvement}. The results suggest that \SAEs are useful augmentation samples as they do not harm generalization performance, as noted by the results in the benign column, but improve performance on unseen \SAEs. 

\begin{table}[h]
\centering
\begin{tabular}{c c c c c }
\toprule 
\multicolumn{5}{c}{\bf Test} \\
{\bf Train} & benign & si-FGSM & sGD & sCW \\
\midrule
\midrule
\text{benign} & 0.986 & 0.529 & 0.657 & 0.485 \\
\text{si-FGSM} & 0.979 & 0.936 & 0.955 & 0.907 \\
\text{sGD} & 0.978 & 0.939 & 0.946 & 0.928 \\
\text{sCW} & 0.98 & 0.942 & 0.959 & 0.95 \\
\bottomrule
\end{tabular}
\caption{\small {\bf Accuracy metrics of robust networks retrained} with adversarial training using \SAEs generated on the benign model. The rows indicate the method used to generate the retraining \SAEs and the columns indicate the dataset used for evaluation. Observe that retraining on \SAEs improves robustness across datasets, regardless of the initial train dataset (larger values are better). All experiments are carried out on VGG-16.}
\label{tab:accuracy_improvement}
\end{table}
\vspace{-2mm}

\noindent{\bf Information Worth:} Using the formulation proposed in \S~\ref{entropy}, we measure the information worth of all strategies highlighted in Table~\ref{tab:strategies_test}. The results are summarized in Table~\ref{tab:entropy}. Observe that halton sampling using the large range generates the most informative points for augmentation. However, as explained in Appendix~\ref{sampling}, such a strategy is highly sample inefficient. Each sample degrades accuracy with low probability and 5$\times$ the number of sampling trials are needed to obtain an effective counterexample. The next best candidates are obtained using strategies formulated in Appendix~\ref{constructions} \ie sCW and si-FGSM. Note that while the results in Table~\ref{tab:strategies_test} may suggest that augmentation samples that cause the most accuracy degradation are preferred, they do not incorporate their {\em realism} ({\em c.f.} Table~\ref{tab:entropy}). Table~\ref{tab:entropy_resnet} in Appendix~\ref{informativeness_resnet} contains the results for ResNet-50, where si-FGSM generates the most informative samples (suggesting that the hyperparameter choices impact the accuracy degradation vs. realism trade-off and, consequently, the {\em information worth}).
\begin{table}[h]
\centering
\resizebox{\textwidth}{!}{
\begin{tabular}{c c c c c c c c c c}
\toprule 
{\bf Membership} &  None & Halton (large) & Halton (small)  &  Random (large) &  Random (small) & sCW &  sGD   & si-FGSM\\
\midrule
\midrule
Binary & 0.1082 & 1.1125 & 0.6165 & 0.8085 & 0.431 & 0.9565 & 0.7218 & 0.887\\
Fractional & 0.1224 & 1.1158 & 0.5964 & 0.8604 & 0.4742 & 0.9464 & 0.7213 & 0.8783 \\
\bottomrule
\end{tabular}
}
\caption{\small {\bf Information worth} of augmentation samples generated using various strategies. Binary membership uses the model prediction, \ie $\mu_i(x) = \mathds{1}[F(x)=i]$ and fractional membership uses the model confidence, \ie $\mu_i(x)=s(F)(x)_i$. Larger values are better. All experiments are carried out on VGG-16.
\vspace{-5mm}}
\label{tab:entropy}
\end{table}

\subsection{Realism}
\label{realism}

We use both qualitative and quantitative approaches to measure realism of the \SAEs. The analysis was carried out using \SAEs generated from VGG-16; results from ResNet-50 are in Appendix~\ref{expts_resnet}. 

\vspace{1mm}
\noindent{\bf 1. FID \& LPIPS Distance:} To obtain a quantitative measure of realism, we measure: (i) the Fr{\'{e}}chet Inception Distance (FID)~\cite{DBLP:journals/corr/HeuselRUNKH17}, a metric known to correlate with human visual quality\footnote{FID is primarily used to measure output quality of generative models and may not be ideal for our purposes.}, and (ii) the LPIPS distance~\cite{lpips}, another popular metric for perceptual similarity. In both cases, the lower the better. The results are summarized in Table~\ref{tab:realism}. We omit scores for Halton (small) and Random (small) as they do not create informative samples. Each cell contains the average score (per class) when the augmenting samples are compared with the samples they were generated from. The results suggest that across both metrics, the \SAEs are realistic.

\begin{table}[h]
\centering
\resizebox{\textwidth}{!}{
\begin{tabular}{c c c c c c c c c c c c c c}
\toprule 
\multicolumn{14}{c}{\bf Class} \\
{\bf Method} & {\bf Strategy} &   0   &  1   &   2 &   3  & 4 & 5 & 6 & 7  & 8  & 9 & 10  &11 \\
\midrule
\midrule
& si-FGSM & 5.62 & 20.56 & 23.88 & 13.40 & 8.56 & 16.25 & 21.14 & 11.05 & 18.30 & 16.91 & 18.68 & 8.04\\
& sGD & 4.33 & 18.73 & 22.97 & 10.99 & 6.20 & 14.19 & 25.53 & 9.33 & 17.17 & 16.79 & 15.39 & 7.45\\
FID & sCW & 6.23 & 26.48 & 29.74 & 11.18 & 8.39 & 15.85 & 31.97 & 11.75 & 21.65 & 17.08 & 15.93 & 7.79\\
& Halton (large) & 2.48 & 32.68 & 21.10 & 15.42 & 13.98 & 9.15 & 38.27 & 5.09 & 21.94 & 27.71 & 20.38 & 5.82\\
& Random (large) & 6.79 & 27.41 & 26.33 & 17.19 & 12.28 & 16.42 & 24.30 & 12.53 & 23.92 & 22.66 & 19.63 & 9.49\\
\midrule
& si-FGSM & 0.25 & 0.53 & 0.40 & 0.60 & 0.45 & 0.43 & 0.36 & 0.41 & 0.44 & 0.33 & 0.47 & 0.37\\
& sGD & 0.22 & 0.52 & 0.37 & 0.58 & 0.41 & 0.42 & 0.37 & 0.38 & 0.46 & 0.34 & 0.45 & 0.35\\
LPIPS & sCW & 0.27 & 0.58 & 0.45 & 0.60 & 0.46 & 0.45 & 0.42 & 0.40 & 0.50 & 0.35 & 0.45 & 0.39\\
& Halton (large) & 0.12 & 0.59 & 0.37 & 0.66 & 0.55 & 0.26 & 0.51 & 0.22 & 0.51 & 0.46 & 0.47 & 0.28\\
& Random (large) & 0.29 & 0.58 & 0.44 & 0.67 & 0.51 & 0.46 & 0.45 & 0.45 & 0.53 & 0.41 & 0.53 & 0.41\\
\bottomrule
\end{tabular}
}
\caption{\small {\bf Realism measures} for augmentation samples generated using VGG-16. Lower the scores, more realistic the images.}
\label{tab:realism}
\end{table}
\vspace{-2mm}



\vspace{1mm}
\noindent{\bf 2. Survey:} To validate if (i) humans are able to correctly classify the objects in \SAEs, and (ii) if humans consider the \SAEs realistic, we conducted out an online survey on Amazon Mechanical Turk with 30 participants\footnote{Each participant was a master worker and was compensated \$8 for the study, approved by our IRB.}. Each participant was asked 2 questions about 15 \SAEs. The first question was to classify the object. The second question was to rate the realism of the modification induced (in comparison to the original, unmodified image placed next to it) on a scale of 1 (lowest) to 10 (highest). For the classification task, we observe that the human participants are able to correctly classify the object 98\% of the time (on average). The average median realism for samples generated using Redner was 6.67 ({\em mean realism} = 6.35). These results further validate the realism of \SAEs.


\subsection{Robustness}
\label{robustness}

Results in Table~\ref{tab:accuracy_improvement} suggest that \SAEs are useful for augmentation. To measure if the retrained models are more robust, we test these models using methods formalized in \S~\ref{map} (and Appendix~\ref{constructions}). Each row in Table~\ref{tab:robustness_improvements} contains normalized accuracy by evaluating the augmented model (as in Table~\ref{tab:strategies_test}) with test samples generated using the augmented model (as opposed to the benign model as in Table~\ref{tab:accuracy_improvement}); this difference is highlighted by the subscript $robust$. In comparison to the benign model (in row 1), we can see that the augmented models are indeed more robust to all adversarial methods tested, regardless of the technique used to generate training \SAEs. This hints at potential cross-method robustness transferability -- \ie training the network to be robust against one \SAE-generation method greatly helps with robustness against other \SAE-generation methods as well.

\begin{table}[h]
\centering
\begin{tabular}{c c c c c}
\toprule 
\multicolumn{5}{c}{\bf Test} \\
{\bf Train} & \text{benign} & \text{si-FGSM$_{robust}$} & \text{sGD$_{robust}$} & \text{sCW$_{robust}$} \\
\midrule
\midrule
\text{benign} & 0.991 & 0.547 & 0.685 & 0.483 \\
\text{si-FGSM} & 0.978 & 0.797 & 0.871 & 0.819 \\
\text{sGD} & 0.978 & 0.836 & 0.910 & 0.797 \\
\text{sCW} & 0.991 & 0.884 & 0.918 & 0.879 \\
\bottomrule
\end{tabular}
\caption{\small {\bf Accuracy metrics for benign and adversarially retrained networks} on \SAEs generated by using various methods (in the columns) on individual networks trained to be robust to individual methods (in the rows). Retraining against one method helps provide some robustness against {\em all} tested methods.}
\label{tab:robustness_improvements}
\end{table}
\vspace{-5mm}

\subsection{Cross-Model Transferability}
\label{transferability}

From Table~\ref{tab:transfer_vgg_to_resnet}, we observe that the \SAEs generated using VGG-16 as the base model transfer to a ResNet-50 model trained on the benign dataset, though to a lesser degree ({\em c.f.} to Table~\ref{tab:strategies_test} for a comparison). This suggests that the ResNet-50 model could be retrained using \SAEs generated using a different model. This can potentially be used to reduce the run-time efficiency of generating \SAEs (as the cost of generating a \SAE is proportional to the depth of the victim network). We validate that \SAEs generated using ResNet-50 transfer over to VGG-16 in Appendix~\ref{resnet_transferability}.

\begin{table}[h]
\centering
\resizebox{\textwidth}{!}{
\begin{tabular}{c  c c c c c c c c c c c c c}
\toprule 
\multicolumn{14}{c}{\bf Class} \\
{\bf Strategy} &   0   &  1   &  2 & 3  & 4    & 5 & 6     &7 & 8 & 9 & 10  & 11   & \textbf{overall}  \\
\midrule
\midrule
\text{sCW}  & 0.998& 	0.713& 	0.621& 	0.858& 	0.451& 	0.916& 	1& 	0.917& 	0.102& 	0& 	0.030& 	0.085& 	{\bf 0.588}  \\
\text{si-FGSM}&  0.998& 	0.786& 	0.857& 	0.834& 	0.495& 	0.933& 	1& 	0.917& 	0.142& 	0.04& 	0.030& 	0.054& 	{\bf 0.621} \\
\text{sGD}&   1& 	0.879& 	0.857& 	0.913& 	0.670& 	0.933& 	1& 	0.945& 	0.122& 	0& 	0.020& 	0.133& 	{\bf 0.716}  \\
\bottomrule
\end{tabular}
}
\caption{\small {\bf Transferability} of \SAEs generated using VGG-16 to ResNet-50. This suggests that images generated using VGG-16 transfer to a ResNet-50 architecture.}
\label{tab:transfer_vgg_to_resnet}
\end{table}
\vspace{-5mm}

\section{Conclusions}
In this paper, we introduce the notion of semantic counterexamples, which are samples that are natural transformations of a sample and which adversely affect an ML model. We demonstrate an approach for generating semantic counterexamples that adapts and combines two key techniques: algorithms from adversarial ML and differentiable rendering. We formulate a metric to verify the informativeness of these counterexamples and further confirm that they help improve the robustness of ML networks when added to the training set.
Future directions include testing the operation of these semantically robust ML networks within the control loop of a CPS, and
making our algorithms aware of the rest of the control loop. 

\subsection*{Acknowledgments}
The authors would like to thank Tzu-Ming Harry Hsu and Tzu-Mao Li for their assistance with setting up the differentiable rendering frameworks. This work was supported in part by NSF FMitF grant 1837132, NSF CPS grant 1545126 (VeHICaL), NSF CCF-FMitF-1836978, NSF SaTC-Frontiers-1804648, NSF CCF-1652140, the DARPA Assured Autonomy project, Berkeley Deep Drive, the iCyPhy center, Air Force Grant FA9550-18-1-0166, and ARO grant number W911NF-17-1-0405.

\bibliographystyle{unsrt}
\bibliography{references}
\newpage
\appendix
\section*{Appendix}
\section{Related Work}
\label{app:related}

Our work builds upon the literature on differentiable rendering and adversarial ML, which we review below.

{\em Differentiable Rendering:} The process of finding 3D scene parameters (geometric, textural, lighting, etc.) given images is referred to as de-rendering or inverse graphics~\cite{Baumgart, Faces, AutoLighting, Shape}. Pipelines for differentiable rendering (the opposite of inverse graphics) were proposed by Loper \etal~\cite{Loper} and Kato \etal~\cite{Kato}.
Kulkarni \etal~\cite{kulkarni2015deep} combine both de-rendering and rendering, and propose a model that learns interpretable representations of images (similar to image semantics), and show how these interpretations can be modified to produce changes in the input space. Li \etal~\cite{li2018differentiable} design a general-purpose differentiable ray tracer; gradients can be computed with respect to arbitrary semantic parameters such as camera pose, scene geometry, materials, and lighting parameters. Similarly, Yao \etal~\cite{3dsdn2018} propose a pipeline that, through de-rendering obtains various forms of semantics, geometry, texture, and appearance, which can be rendered using a generative model. Our experimental setup uses both of these differentiable renderers. 


{\em Adversarial ML:} Adversarial examples are test-time inputs that result in incorrect outputs produced by the ML model. Extensive prior work~\cite{Madry,FGSM,carlini2017towards,moosavi2016deepfool,papernot2016limitations} focuses on generating norm-bounded pixel-level changes to input features to induce incorrect actions. However, these manipulations are not usually realizable in the real world. 
Our focus is on those attacks that result in {\em semantically meaningful} adversarial samples. 
Generating real world adversarial examples has resulted in several efforts (e.g.,~\cite{athalye2017synthesizing,eykholt2018robust,bhattad2020unrestricted}), some of which are more realistic than others. 
Eykholt \etal~\cite{eykholt2018robust} show how physical changes can fool neural networks.
Bhattad \etal~\cite{bhattad2020unrestricted} give techniques for colorizing or changing the
texture of images to generate adversarial examples.
Geirhos \etal~\cite{Texture} discovered that certain models are biased towards textural cues, while Engstrom \etal~\cite{engstrom2019exploring} observe that modifying the spatial orientation of images results in misclassifications. Sharif \etal~\cite{sharif2016accessorize} create semantic attacks against facial recognition systems by {\em realizing} changes in pixel intensities by placing brightly colored objects on faces. Song \etal \cite{songGAN} develop an approach for creating semantic adversarial examples, but utilize GANs to do this instead, using class-conditional searches on the latent space. This work, however, constructs examples from scratch instead of taking existing datapoints and perturbing them, and does not utilize gradient-based approaches, limiting the flexibility of their pipeline. Meanwhile, approaches that involve explicitly sampling the semantic parameter space are slow and expensive~\cite{SAESeshia}. The work that is closest to ours is that of Xiao \etal~\cite{Xiao_2019_CVPR} and Qiu \etal~\cite{SemanticAdv}; the former uses a differentiable renderer to induce changes in shapes and textures, while the latter induces a different set of semantic changes using a generative model. 
Our approach stands out by being general, in that it can apply to any vector of semantic features as long as we can render from those features using a differentiable renderer. Further, we show how our generated semantic counterexamples can be effectively used for data augmentation and that they can transfer between DNNs.

\section{Constructions}
\label{constructions}

\paragraph{1. Semantic iterative FGSM.}
Recall that $z=(x,y)$ where $x = \render(\seminp)$ and $y=F(x)=F(\render(\seminp))$.
Consider the loss function $\ell_F(x,y)$. Let $\seminp_0$ be a starting semantic feature vector that we wish to perturb
into a semantic counterexample.
The derivative with respect to $\seminp$, evaluated at $\seminp_0$ is $\left.\left[\frac{\partial\render}{\partial\seminp}\right]^{\top}\right\vert_{\seminp_0}\left.\nabla_{z} \ell_F (z)\right\vert_{z = (\render(\seminp_0), F(\render(\seminp_0)))}$ (the notation $\left.\left[\frac{\partial\render}{\partial\seminp}\right]^{\top}\right\vert_{\seminp_0}$ is the transposed Jacobian matrix of $\render$ as a vector-valued function of $\seminp$, evaluated at $\seminp_0$, and  $\nabla_{z} \ell_F(z) |_{z = (\render(\seminp_0), F(\render(\seminp_0)))}$ is the derivative evaluated at 
$(\render(\seminp_0), F(\render(\seminp_0)))$.

The semantic version of FGSM (\ie sFGSM) will, given the step-size hyperparameter $\alpha$, produce the following semantic counterexample $\semce$:
\begin{equation}
\label{eq:sFGSM}
\semce  = \seminp_0 + \alpha \signf\left(
\left.\left[\tfrac{\partial\render}{\partial\seminp}\right]^{\top}\right\vert_{\seminp_0}\left.\nabla_{z} \ell_F (z)\right\vert_{z = (\render(\seminp_0), F(\render(\seminp_0)))}
%
%
\right)
\end{equation}
Note that we only assume that the rendering function $\render$ is differentiable.


In a similar manner a semantic version of the iterative FGSM attack ({si-FGSM}) can be constructed.  Let $\seminp_0$ be the initial semantic feature vector. The update steps correspond to the following equation:

\begin{equation}
\label{eq:iFGSM}
\seminp_{k+1}  = \seminp_k + \alpha \signf\left(
\left.\left[\tfrac{\partial\render}{\partial\seminp}\right]^{\top}\right\vert_{\seminp_k}\left.\nabla_{z} \ell_F (z)\right\vert_{z = (\render(\seminp_k), F(\render(\seminp_k)))}
\right)
\end{equation}

Similar to Simonyan \etal~\cite{simonyan2013deep}, we define $\ell_F$ to be the the raw (unnormalized) class score of the correct label instead of the normalized softmax probability for the attack listed above.

\paragraph{2. Semantic GD.} We present a semantic version of the GD attack ({sGD}). Again, let $\seminp_0$ be the initial semantic feature vector. The update steps in semantic GD correspond to the following equation:

\begin{equation}
\label{eq:GD}
\seminp_{k+1}  = \Pi_{B_\nu (\seminp_0,\varepsilon)} \left( \seminp_k + \alpha \cdot \left(
\left.\left[\tfrac{\partial\render}{\partial\seminp}\right]^{\top}\right\vert_{\seminp_k}\left.\nabla_{z} \ell_F (z)\right\vert_{z = (\render(\seminp_k), F(\render(\seminp_k)))}
\right)
\right)
\end{equation}

Similar to Simonyan \etal~\cite{simonyan2013deep}, we define $\ell_F$ to be the the raw (unnormalized) class score of the correct label instead of the normalized softmax probability for the attack listed above.

Note that $\Pi_{B_\nu (\cdot,\cdot)}$ is the projection operator in the parameter space $\Theta$. We also assume that the 
projection operator will keep the parameters in the feasible 
set, which depends on the image (\eg translation does not take the 
car off the road).

\paragraph{3. Semantic CW.} The semantic Carlini-Wagner (sCW) attack solves the following optimization problem.
\begin{equation}
    \min\|\semmod\|_p + c \cdot f(\seminp + \semmod)
\end{equation}
where $p=1$, $c=0.1$ and the objective function $f$ is defined as,
\begin{equation}
    f(\theta + \pi) = \max((\max_{i\ne t} \Phi_i(R(\theta + \pi)) - \Phi_t(R(\theta+ \pi))), 0)
\end{equation}
and $\Phi_i$ is the output corresponding to class $i$ of the model $F$ before the softmax layer. The gradients are computed using the Adam optimizer~\cite{kingma2014adam}, with the learning rate $\eta$.



\section{Notes On Augmentation Quality}
\label{notes_augmentation}

\subsection{Defining Information Worth}
\label{app:entropy}

Based on the notation formalized in \S~\ref{entropy}, let $E_i$ be the entropy measured on each subset $D_i$ of size $m_i$ \ie $E_i = -\sum_{j\in L} p_j^i\log p_j^i$, where $p_j^i$ is the estimated probability of observing the class $j$ in the subset $D_i$.


\begin{align*}
    p_j^i = \frac{\sum_{t=1}^{m_i} \mathds{1}[x_t\in D_i] \mathds{1}[y_t=j]}{m_i} = \frac{\sum_{t=1}^{m_i} \mathds{1}[f(x_t)=i] \mathds{1}[y_t=j]}{m_i}
\end{align*}

If the entropy $E_i$ for each subset $D_i$ is low, then the classifier divides $D$ in a "well-organized" way, \ie the subset is extremely likely to consist of datapoints of single label. Thus, the following weighted average of $E_i$'s (henceforth referred to as {\em information worth}) is used to estimated the quality of a classifier \ie  $\hat E = \sum_{i\in[N]}\alpha_i \cdot E_i$, where each weight $\alpha_i$ is determined by the proportion of the size of the subset $D_i$ to the size of the entire set $D$ \ie $\alpha_i = \frac{|D_i|}{|D|} = \frac{m_i}{m}$. 

\subsection{Salient Features of Information Worth}

We make the following observations:
\begin{itemize}
\itemsep0em
\item A classifier achieving high accuracy is likely to achieve low value of $\hat E$ (however, the converse is not true in general). Therefore, a high value $\hat E$ can be an indicator of a bad classifier.
\item If we update a classifier (through training) from $F$ to $F'$, we can measure the quality of the change in terms of the {\em information gain} which is the amount of decrease $\hat E_F - \hat E_{F'}$ (larger, the better).
\end{itemize}

\section{Pipelines}
\label{app:pipelines}

\subsection{Overall Pipeline For Generating Semantic Counterexamples For Classification}

\begin{figure}[H]
\centering
\includegraphics[width=0.8\textwidth]{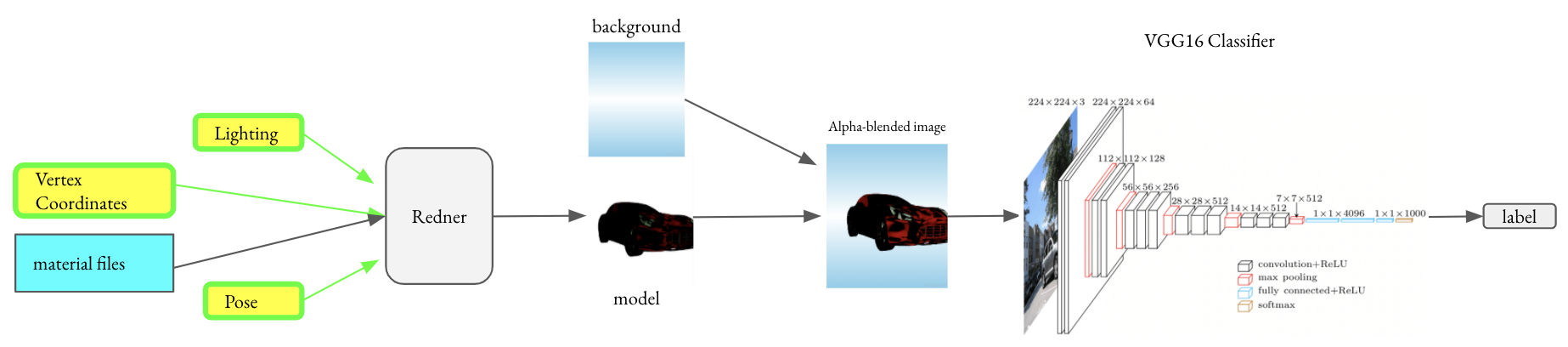}
\caption{\small Overall pipeline for generating \SAEs for classification. The renderer (Redner) is fed vertex, lighting, material, and pose information. The output model is then alpha-blended with a background and the resulting image is classified a DNN (say VGG-16~\cite{vgg}). The gradients of the output are computed with respect to the semantic parameters (in yellow with green arrows). They can then be adversarially perturbed at the points indicated by the arrows highlighted in green and the resulting image is re-rendered and re-classified. We borrow the graphic for VGG-16 from~\cite{VGG16graphic}.}
\label{fig:redner_pipeline}
\end{figure}

\subsection{Overall Pipeline For Generating Semantic Counterexamples For Detection   }

\begin{figure}[H]
\centering
\includegraphics[width=0.8\textwidth]{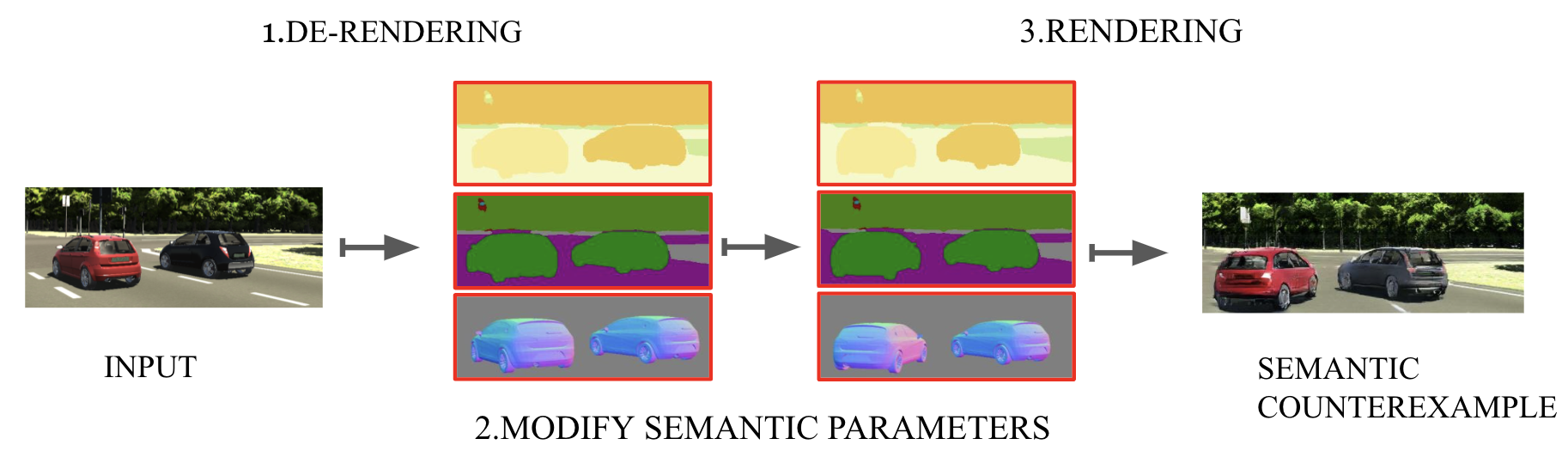}
\caption{\small Our procedure for attacking an object detection framework: The input is de-rendered (step 1) to its intermediary representation (IR) - semantic, graphic, and textural maps. Then, this is adversarially perturbed (\eg the red car is rotated) as described in \S~\ref{sec:sae} (step 2). The resulting IR is then re-rendered to the generate the counterexample (step 3).}
\label{fig:3dsdn_pipeline}
\end{figure}
\section{Implementation Specifics} 
\label{specifics_implementation}

\subsection{Semantic Features}
\label{semantic}

Recall that we generate \SAEs by strategically perturbing semantic features. 

In the case of VKITTI (using 3D-SDN), the semantic features include (i) \cars: the car's texture codes which change its color, (ii) \sky: the weather and time of day, (iii) \foliage: the surrounding foliage and scenery, (iv) \rotate: the car's orientation, (v) \translate: the car's position in 2D, and (vi) \FFD: the 3D mesh which provides structure to the car. 

In the case of ShapeNet (using Redner), the semantic features include: (i) \pose: the camera angle at which the object is viewed in the frame, (ii) \vertex: the vertex coordinates of the object, and (iii) \lighting: the environment lighting of the scene

\subsection{Semantic Parameter Combinations}
\label{combinations}

Before we discuss specifics about the strategies we use to generate \SAEs, we need to understand how the semantic parameter space can be manipulated. In particular, we wish to answer the question: {\em Which semantic parameters do we change?} In the pixel perturbation setting, all pixels are equal \ie any pixel can be perturbed. Whether such uniformity naturally exists in the semantic space is unclear. However, we have additional flexibility; for each rendering framework, we can choose to modify any of the above listed semantic parameters independently without altering the others, \ie perform {\em single parameter modifications}. Alternatively, we can modify any subset of the parameters in unison, \ie perform {\em multi-parameter modifications}. 

\noindent{\bf Case: 3D-SDN + SqueezeDet + VKITTI}

Semantic counterexamples are generated using an iterative variant of FGSM (\ie si-FGSM). 

\begin{table}[h]
\centering
\begin{tabular}{c c c c c c c}
\toprule 
\multicolumn{6}{c}{\bf Parameters} \\
\textbf{Metric}  &   \cars   &   \sky   &   \foliage     &   \translate  &   \rotate     &   \FFD    \\
\midrule
\midrule
\textbf{recall}  &   100     &   100      &   100          &   100         &   100         &   98.7 \\ 
\textbf{mAP}        &   99.5     &   98.8       &   99.7      &   99.2        &   98.2        &   98.7   \\   
\bottomrule
\end{tabular}
\caption{\small  Ineffectiveness of single parameter modifications. Performance of SqueezeDet on SAEs generated using single parameter modifications. The model had (a) recall = 100, and (b) mAP = 99.4 on benign/non-adversarial inputs.}
\label{tab:single}
\end{table}

For multi-parameter modifications, we consider the following combination of semantic parameters: (a) \texttt{combo 1} = \translate + \rotate, (b) \texttt{combo 2} = \translate + \rotate + \FFD, (c) \texttt{combo 3} = \translate + \FFD, and (d) \texttt{combo 4} = \rotate + \FFD. The results are reported in Table~\ref{tab:multi}. We observe that for \texttt{combo 2}, there is $\sim 35$ PPs decrease in mAP. \texttt{combo 2} will be used to generate \SAEs through the remainder of the paper.

\begin{table}[h]
\centering
\begin{tabular}{c  c c c c c}
\toprule 
\multicolumn{6}{c}{\bf Parameters} \\
{\bf Metric}  &   Baseline & \texttt{combo 1}   &   \texttt{combo 2}   &   \texttt{combo 3}        &   \texttt{combo 4}  \\
\midrule
\midrule
\textbf{recall}  &   100 & 100      &   100        &   100         & 100   \\   
\textbf{mAP}        & 99.4 &   82       &   65.9       &   80.8        &  98.7    \\   
\bottomrule
\end{tabular}
\caption{\small  Effectiveness of multi-parameter modifications for 3D-SDN.}
\label{tab:multi}
\end{table}
\vspace{-1mm}

\noindent{\bf Case: Redner + VGG-16 + ShapeNet} 

As in the previous case, single parameter modifications are not as effective as multi-parameter modifications (refer Table~\ref{tab:redner_multi} for normalized accuracy measurements). Additionally, modifications to \pose induce greater accuracy degradation than modifications to \vertex. For all results reported in \S~\ref{sec:eval}, we use \SAEs obtained through multi-parameter modifications. Note that changing the \lighting parameter did not impact the accuracy of classification, and is consequently omitted from all tabulations.

\begin{table}[!h]
\centering
\begin{tabular}{c c c c}
\toprule 
\multicolumn{4}{c}{\bf Parameters} \\
\textbf{Strategy} &  \text{\vertex} & \text{\pose} & \text{\pose + \vertex}  \\
\midrule
\midrule
\text{sCW} & 0.892 & 0.543 & 0.485   \\
\text{si-FGSM}  &  0.862 & 0.615 & 0.529 \\
\text{sGD} & 0.897 & 0.741 & 0.657 \\
\bottomrule
\end{tabular}
\caption{\small  Effectiveness of multi-parameter modifications for Redner.}
\label{tab:redner_multi}
\end{table}

\subsection{Hyperparameters For Generating Semantic Counterexamples}

In the context of generating \SAEs, the value of the step-size parameter $\alpha$ is proportional to the magnitude of the geometric and textural changes induced; the effect depends on the semantic parameter under consideration. Large values of $\alpha$ result in unrealistic images rendered. To avoid such issues and to simulate {\em realistic} transformations, we use a different step-size for each semantic parameter. Similarly, the projection bound $\varepsilon$ enables us to clip perturbations that exceed this bound (refer to the construction of PGD~\cite{Madry} and sGD for its use). 

\noindent{\bf 1. Redner:} Below, we report the hypeparameters we use for the various strategies formalized in Appendix~\ref{constructions}. These are specific to generating \SAEs using VGG-16 (refer Table~\ref{tab:params_vgg}). For sCW, the attack was formulated according based on the work of Carlini \etal~\cite{carlini2017towards} with the 2-norm used for our distance metric. The attack was untargeted, as we sought just to induce a misclassification of some kind while minimizing the 2-norm of the distance between the old semantic feature vector and the new one.

\begin{table}[ht]
\centering
\begin{tabular}{c c c c c c c c}
\toprule 
\textbf{Strategy} &  $\alpha_{vertex}$ & $\alpha_{pose}$ & \text{\# iterations} & $\varepsilon_{vertex}$ & $\varepsilon_{pose}$ & $\eta_{vertex}$ & $\eta_{pose}$\\
\\
\midrule
\midrule
\text{sCW} & - & - & 5 & - & - & 0.01 & 0.30 \\
\text{si-FGSM}  &  0.002 & 0.15 & 5  & - & - & - & -\\
\text{sGD} & 0.01 & 0.20 & 5 & 0.05 & 1.0 & - & -\\
\bottomrule
\end{tabular}
\caption{\small  Hyperaparameters used for generating \SAEs using VGG-16. $\alpha$ is step-size, $\varepsilon$ is projection bound, $\eta$ is the learning rate}
\label{tab:params_vgg}
\end{table}

These hyperparameters specified below are specific to generating \SAEs using ResNet-50 (refer Table~\ref{tab:params_resnet}).

\begin{table}[!h]
\centering
\begin{tabular}{c c c c c c c c}
\toprule 
\textbf{Strategy} &  $\alpha_{vertex}$ & $\alpha_{pose}$ & \text{\# iterations} & $\varepsilon_{vertex}$ & $\varepsilon_{pose}$ & $\eta_{vertex}$ & $\eta_{pose}$\\
\midrule
\midrule
\text{sCW} & - & - & 5 & - & - & 0.01 & 0.90\\
\text{si-FGSM}  &  0.01 & 0.45 & 5  & - & - & - & - \\
\text{sGD} & 0.01 & 0.60 & 5 & 0.05 & 3.0 & - & - \\
\bottomrule
\end{tabular}
\caption{\small  Hyperparameters used for generating \SAEs using ResNet-50. $\alpha$ is step-size, $\varepsilon$ is projection bound, $\eta$ is the learning rate}
\label{tab:params_resnet}
\end{table}

\noindent{\bf 2. 3D-SDN:} We utilize an iterative variant of FGSM to modify the semantic parameters (\ie si-FGSM). Recall that in FGSM, the degree of modification of the input is determined by the step-size hyperparameter $\alpha \in [0,1]$. The hyperparameters are specified in Table~\ref{tab:params_squeezedet}. 

\begin{table}[h]
\centering
\begin{tabular}{c c c c c c c}
\toprule 
\multicolumn{6}{c}{\bf Parameters} \\
\textbf{}  &   \cars   &   \sky   &   \foliage     &   \translate  &   \rotate     &   \FFD    \\
\midrule
\midrule
$\alpha$ & 0.05 & 0.25 & 0.10 & 0.01 & 0.01 & 0.025 \\   
\bottomrule
\end{tabular}
\caption{\small  Hyperparameters for generating \SAEs using SqueezeDet. $\alpha$ is the step-size.}
\label{tab:params_squeezedet}
\end{table}

We stress that for both differentiable graphics frameworks, these hyperparameters were obtained after extensive visual inspection (by 3 viewers independently). An added benefit of our particular choice of hyperparameters (for 3D-SDN) is that it enables us to use the same ground truth labels throughout our experiments; the produced \SAEs have bounding box coordinates that enable us to use the same ground truth labels as their benign counterparts\footnote{This fact is useful when we evaluate model robustness through retraining the models with \SAEs as inputs.}. 

\subsection{Datasets}
\label{datasets}

Table~\ref{tab:dataset_redner} contains salient features of the ShapeNet dataset we use for experiments with Redner. There are a total of 3670 test images, 12843 train images, and 1835 validation images. Unless specified otherwise, this proportion is used for all experiments. The class imbalance stems from the lack of availability of samples for specific classes in ShapeNet. The experiments in \S~\ref{robustness} use a smaller test set of 232 images. that roughly maintains the proportions present in the full test dataset.  
\begin{table}[!h]
\centering
\begin{tabular}{c c c c c}
\toprule 
\text{Class} &  \text{Name} & \text{\# test samples} & \text{\# train samples} & \text{\# validation samples} \\
\midrule
\midrule
0 & airplane & 775 & 2713 & 388 \\
1 & bench & 464 & 1624 & 232\\
2 & trashcan & 119 & 417 & 60\\
3 & bus & 127 & 445 & 64\\
4 & car & 1549 & 5421 & 774 \\
5 & helmet & 60 & 210 & 30 \\
6 & mailbox & 15 & 53 & 8 \\
7 & motorcycle & 73 & 258 & 37 \\
8 & skateboard & 49 & 171 & 24 \\
9 & tower & 25 & 87 & 12 \\
10 & train & 99 & 344 & 49 \\
11 & boat & 315 & 1100 & 157 \\
\bottomrule
\end{tabular}
\caption{\small  Salient features of the ShapeNet dataset.}
\label{tab:dataset_redner}
\end{table}

For experiments with VKITTI, our train dataset consisted of 6339 images, and the test dataset consisted of 2082 images. Each image comprised of the following objects: (a) one or more cars, (b) one or more buses, (c) {\em no pedestrians}. The object detection task involved detecting the car(s) and/or bus(es).

\subsection{Sampling Approaches}
\label{sampling}

We utilize sampling based approaches as baselines only for the experiments involving image classification \ie ShapeNet and VGG-16/ResNet-50 using Redner.

\noindent{\bf 1. Random Sampling:} For each benign point, we obtain 5 samples using Halton sampling in 2 distinct parameter ranges (only for \pose transformations): (i) large \ie [-0.75, +0.75] radians, and (ii) small [-0.3, +0.3] radians. For each range, among the 5 samples generated, we first verify if the sample induces a misclassification. If so, we pick the one with the highest softmax value for the incorrect prediction.

\noindent{\bf 2. Halton Sampling:} We utilize Halton sampling as implemented by Dreossi \etal~\cite{dreossi-ijcai18}; for each benign point, we obtain 5 samples using Halton sampling in 2 distinct parameter ranges (only for \pose transformations): (i) large \ie [-0.75, +0.75] radians, and (ii) small [-0.3, +0.3] radians. For each range, among the 5 samples generated, we first verify if the sample induces a misclassification. If so, we pick the one with the highest softmax value for the incorrect prediction.

Observe that such passive sampling approaches are sample inefficient. To obtain one counterexample needed for augmentation, we are required to generate $N=5$ samples and obtain their corresponding softmax values (by running inference on all samples). If one were to randomly pick a sample generated by either sampling strategy, there is no guarantee that the point is informative and can be used for augmentation. Active sampling strategies are more inefficient. In contrast, the mechanisms detailed in \S~\ref{map} generate a \SAE that is optimized to be informative.

\subsection{Training Procedure}

We detail the procedure we used for training to obtain both the benign and robust models.

\noindent{\bf 1. Image Classification:} A pretrained ImageNet~\cite{deng2009imagenet} model is taken and the final fully connected layer is retrained in both VGG-16 and ResNet-50, for 20 epochs, using the images in Table~\ref{tab:dataset_redner}, to obtain the benign models we use in \S~\ref{sec:eval}. To obtain the robust models, we replace half of the training set (at random) with the corresponding \SAEs, and retrain the benign model for 20 epochs. 

\noindent{\bf 2. Object Detection:} The differentiable graphics framework induces several artifacts on rendering. To this end, we first used identity transforms (\ie passed the benign image through the differentiable graphics framework without semantically modifying it) to obtain benign re-rendered images. The entire pretrained SqueezeDet model is retrained for 24000 steps using all the 6339 re-rendered images to obtain the benign model. To obtain the robust models, we replace a quarter of the training set at random with \SAEs, and retrain the benign model for 6000 steps. 
\section{ResNet-50 Experiments}
\label{expts_resnet}

\subsection{Informativeness}
\label{informativeness_resnet}

We repeat the same experiments as in \S~\ref{gain}. While the results broadly follow a similar trend, observe that \SAEs are more informative in comparison to samples generated using Halton sampling or random sampling (refer Table~\ref{tab:strategies_test_resnet} and Table~\ref{tab:entropy_resnet}). 
\begin{table}[h]
\centering
\resizebox{\textwidth}{!}{
\begin{tabular}{c  c c c c c c c c c c c c c}
\toprule 
\multicolumn{14}{c}{\bf Class} \\
{\bf Strategy} &   0   &  1   &  2 & 3  & 4    & 5 & 6     &7 & 8 & 9 & 10  & 11   & \textbf{overall}  \\
\midrule
\midrule
\text{benign}  &  1	& 1	& 1	& 1& 0.998	& 1& 	1& 	1& 	1& 	1& 	0.989& 	0.984& {\bf 0.997} \\
\text{si-FGSM} & 0.993 & 0.560 & 0.512 & 0.677 & 0.267 & 0.85 & 0.666 & 0.821 & 0 & 0 & 0 & 0.025 & {\bf 0.468}   \\
\text{sCW} & 0.997	& 0.849& 	0.874& 	0.740& 	0.683& 	0.916& 	0.666& 	0.863& 	0&	0& 	0.010& 	0.025& 	{\bf 0.697} \\
\text{sGD}  &  0.998&  	0.760&  	0.680&  	0.811&  	0.51&  	0.966&  	0.733&  	0.904&  	0.020&  	0&  	0&  	0.031&  	{\bf 0.612}   \\
\text{Random (large)} & 0.983& 	0.687& 	0.571& 	0.685& 	0.345& 	0.916& 	0.6& 	0.739& 	0& 	0.04& 	0.010& 	0.057& 	{\bf 0.520}\\
\text{Random (small)} &  1& 	0.963& 	0.916& 	0.818& 	0.715& 	1& 	1& 	0.945& 	0.122& 	0.04& 	0.020& 	0.174& 	{\bf 0.749}\\
\text{Halton (large)} & 0.984& 	0.553& 	0.226& 	0.629& 	0.052& 	0.933& 	0.466& 	0.863& 	0.081& 	0& 	0& 	0.212& 	{\bf 0.383}\\
\text{Halton (small)} & 1& 	0.982& 	0.932& 	0.897& 	0.757& 	1& 	1& 	1& 	0.571& 	0.12& 	0.030& 	0.317& 	{\bf 0.793}\\
\bottomrule
\end{tabular}
}
\caption{\small Accuracy degradation induced by different augmentation strategies. Observe that \SAEs are effective at inducing accuracy degradation (lower the {\bf overall} value, the better). All experiments are carried out using ResNet-50 as the target model.}
\label{tab:strategies_test_resnet}
\end{table}

\begin{table}[h]
\centering
\begin{tabular}{c c c c c }
\toprule 
\multicolumn{5}{c}{\bf Test} \\
{\bf Train} & benign & si-FGSM & sGD & sCW \\
\midrule
\midrule
\text{benign} & 0.986 & 0.469 & 0.612 & 0.698 \\
\text{si-FGSM} & 0.978 & 0.946 & 0.938 & 0.941 \\
\text{sGD} & 0.977 & 0.914 & 0.947 & 0.940 \\
\text{sCW} & 0.984 & 0.891 & 0.925 & 0.955 \\
\bottomrule
\end{tabular}
\caption{\small Accuracy metrics of robust networks retrained with adversarial training using \SAEs generated on the benign model. The rows indicate the attack method used to generate the retraining \SAEs and the columns indicate the attack dataset used for evaluation. Observe that retraining on \SAEs improves robustness across datasets, regardless of the initial train dataset (larger values are better). All experiments are carried out on ResNet-50.}
\label{tab:accuracy_improvement_resnet}
\end{table}

\begin{table}[H]
\centering
\resizebox{\textwidth}{!}{
\begin{tabular}{c c c c c c c c c c}
\toprule 
{\bf Membership} &  None & Halton (large) & Halton (small)  &  Random (large) &  Random (small) & sCW &  sGD   & si-FGSM\\
\midrule
\midrule
Binary      & 0.108 & 0.979 & 0.518 & 0.968 & 0.604 & 0.656 & 0.236 & 0.978 \\
Fractional  & 0.122 & 1.029 & 0.548 & 1.017 & 0.657 & 0.697 & 0.258 & 1.033 \\
\bottomrule
\end{tabular}
}
\caption{\small Information worth of augmentation samples generated using various strategies. Larger values are better. Binary membership uses the model prediction, \ie $\mu_i(x) = \mathds{1}[F(x)=i]$ and fractional membership uses the model confidence, \ie $\mu_i(x)=s(F)(x)_i$. Larger values are better. All experiments are carried out on ResNet-50.}
\label{tab:entropy_resnet}
\end{table}
\vspace{-2mm}

\subsection{Realism}

The results are reported in Table~\ref{tab:realism_resnet}.

\begin{table}[H]
\centering
\resizebox{\textwidth}{!}{
\begin{tabular}{c c c c c c c c c c c c c c}
\toprule 
\multicolumn{14}{c}{\bf Class} \\
{\bf Method} & {\bf Strategy} &   0   &  1   &   2 &   3  & 4 & 5 & 6 & 7  & 8  & 9 & 10  &11 \\
\midrule
\midrule
& si-FGSM & 7.99 & 31.97 & 33.87 & 17.62 & 11.02 & 19.15 & 38.60 & 13.35 & 27.11 & 30.10 & 19.98 & 12.46\\
FID & sGD & 5.76 & 29.27 & 32.07 & 14.29 & 7.83 & 16.03 &  40.62 & 11.58 & 26.54 & 27.43 & 20.10 & 11.93\\
& sCW & 2.57 & 6.31 & 6.08 & 12.88 & 3.91 & 15.03 & 19.30 & 12.46 & 27.62 & 29.80 & 19.43 & 12.84\\
\midrule
& si-FGSM & 0.31 & 0.60 & 0.50 & 0.70 & 0.52 & 0.49 & 0.50 & 0.48 & 0.56 & 0.51 & 0.56 & 0.46\\
LPIPS & sGD & 0.26&  0.58&  0.45&  0.67&  0.49&  0.45&  0.50&  0.45&  0.55&  0.49&  0.53&  0.45\\
& sCW & 0.19 & 0.36 & 0.29 & 0.68 & 0.41 & 0.44 & 0.37 & 0.47 & 0.59 & 0.51 & 0.55 & 0.47\\
\bottomrule
\end{tabular}
}
\caption{\small Realism measures for augmentation samples generated using ResNet-50.}
\label{tab:realism_resnet}
\end{table}

\subsection{Cross-Model Transferability}
\label{resnet_transferability}

The results are reported in Table~\ref{tab:transfer_resnet_to_vgg}.

\begin{table}[H]
\centering
\resizebox{\textwidth}{!}{
\begin{tabular}{c  c c c c c c c c c c c c c}
\toprule 
\multicolumn{14}{c}{\bf Class} \\
{\bf Strategy} &   0   &  1   &  2 & 3  & 4    & 5 & 6     &7 & 8 & 9 & 10  & 11   & \textbf{overall}  \\
\midrule
\midrule
\text{sCW}  & 0.991 & 0.803 & 0.848 & 0.212 & 0.785 & 0.9 & 0.533 & 0.616 & 0.020 & 0.04 & 0.272 & 0.542 & \textbf{0.761}  \\
\text{si-FGSM}&  0.878 & 0.521 & 0.521 & 0.078 & 0.362 & 0.916 & 0.2 & 0.685 & 0.040 & 0 & 0.252 & 0.562 & \textbf{0.509} \\
\text{sGD}&   0.943 & 0.618 & 0.495 & 0.252 & 0.635 & 0.983 & 0.066 & 0.822 & 0.082 & 0.04 & 0.353 & 0.619 & \textbf{0.667} \\
\bottomrule
\end{tabular}
}
\caption{\small Transferability of \SAEs generated using ResNet-50 to VGG-16. Lower the {\bf overall} value, the better.}
\label{tab:transfer_resnet_to_vgg}
\end{table}
\vspace{-2mm}

\subsection{Robustness}
\label{resnet_robustness}

\begin{table}[h]
\centering
\begin{tabular}{c c c c c}
\toprule 
\multicolumn{5}{c}{\bf Test} \\
{\bf Train} & \text{benign} & \text{si-FGSM$_{robust}$} & \text{sGD$_{robust}$} & \text{sCW$_{robust}$} \\
\midrule
\midrule
\text{benign} & 0.991 & 0.547 & 0.685 & 0.483 \\
\text{si-FGSM} & 0.978 & 0.5 & 0.331 & 0.698 \\
\text{sGD} & 0.978 & 0.771 & 0.569 & 0.862 \\
\text{sCW} & 0.991 & 0.733 & 0.586 & 0.711 \\
\bottomrule
\end{tabular}
\caption{\small {\bf Accuracy metrics for benign and adversarially retrained networks} on \SAEs generated by using various methods (in the columns) on individual networks trained to be robust to individual methods (in the rows). Retraining against one method helps provide some robustness against {\em all} tested methods.}
\label{tab:robustness_improvements_resnet}
\end{table}
\vspace{-5mm}
\section{3D-SDN Experiments}
\label{3dsdn_expts}

\subsection{Notation}
\label{app:3dsdn_notation}

Let $x$ be an image which contains $N$ recognition targets $T = \{t_1,t_2,...,t_N\}$. Each target $t_n$, $n = 1,2,\cdots,N$, is assigned a ground-truth class label $y_n \in \{1, 2, \cdots , L\}$, where $L$ is the number of classes. Denote by $\mathcal{L} = \{l_1,l_2,...,l_L\}$. The detailed form of $T$ varies among different tasks. In image classification, $T$ contains the whole image. For object detection, $T$ is composed of all pixels. Given a deep network for a specific task, we use $F(x, t_n ) \in \Real^L$ to denote the classification score vector (before softmax normalization) on the $n$-th recognition target of $x$. The exact formulation of the loss functions is dependent on the object detection network's architecture. For example, the loss functions used in SqueezeDet can be found in \S~3.3 of Wu \etal~\cite{squeezedet}.

\subsection{Accuracy Degradation}

We measure the degradation of mean average precision (mAP) and recall on the SqueezeDet object detector~\cite[\S3.3]{squeezedet}. Note that SqueezeDet's loss function comprises three terms corresponding to (a) bounding box regression, (b) confidence score regression, and (c) classification loss. In our experiment, we target the confidence score regression loss term to impact the mAP and recall of the model. Results in Table~\ref{tab:multi} show the efficacy of our multi-parameter augmentation strategy. 

\subsection{Realism}

\noindent{\bf 1. FID and LPIPS scores:} We observe that for the augmenting samples generated using 3D-SDN, the average FID score is 0.102362. The corresponding average LPIPS distance is 0.523521. Observe that the value of the FID score is much lower than those discussed in \S~\ref{realism}.

\noindent{\bf 2. Survey:} We follow the same protocol (and use the same survey) as described in \S~\ref{realism}. For those samples generated using 3D-SDN, the average realism rating was 4.74 (average median = 4.26). In the case of 3D-SDN, the lower scores can be attributed to the poor rendering quality of the differentiable graphics framework; the perturbations made do not alter the geospatial positioning/orientation of the objects in the scene. 

\subsection{Robustness}

When the benign model is retrained to be more robust, we notice that (i) its mAP improves by 7 percentage points (in comparison to Table~\ref{tab:multi}) on \SAEs, and (ii) on benign inputs, the map is only 15 percentage points lesser than the baseline (of mAP=99.4).

\begin{table}[H]
\centering
\begin{tabular}{c c c}
\toprule 
\multicolumn{3}{c}{\bf Test} \\
{\bf Metric}   &  Semantic & Benign  \\
\midrule
\midrule
\textbf{recall}  &  92.97  & 93.7 \\   
\textbf{mAP}     &  72.76  & 84.73 \\   
\bottomrule
\end{tabular}
\caption{\small Robustness improvements due to semantic adversarial training.}
\label{tab:augmentation}
\end{table}

\subsection{Transferability}

To investigate transferability in the object detection task, we train a YOLOv3 network~\cite{redmon2018yolov3} on the KITTI dataset and observe if the \SAEs (generated using SqueezeDet) induce performance degradation. On benign test inputs, the mAP of this network is 91.28\%. When the test set is populated with the SqueezeDet-generated \SAEs, we observed a drop in mAP to 87.50\% (which is not as significant as in the case with SqueezeDet). This suggests that \SAEs (generated for detection) are most effective against the network they are generated from. However, we are not categorically ruling out transferability for the 3D-SDN case (our experiments did not involve testing a variety of hyperparameters as the \SAE generation process using 3D-SDN is very time consuming).

\section{Sample Outputs}
\label{app:samples}

\begin{figure}[!htpb]
\begin{minipage}[b]{0.5\textwidth}
\includegraphics[width=\textwidth]{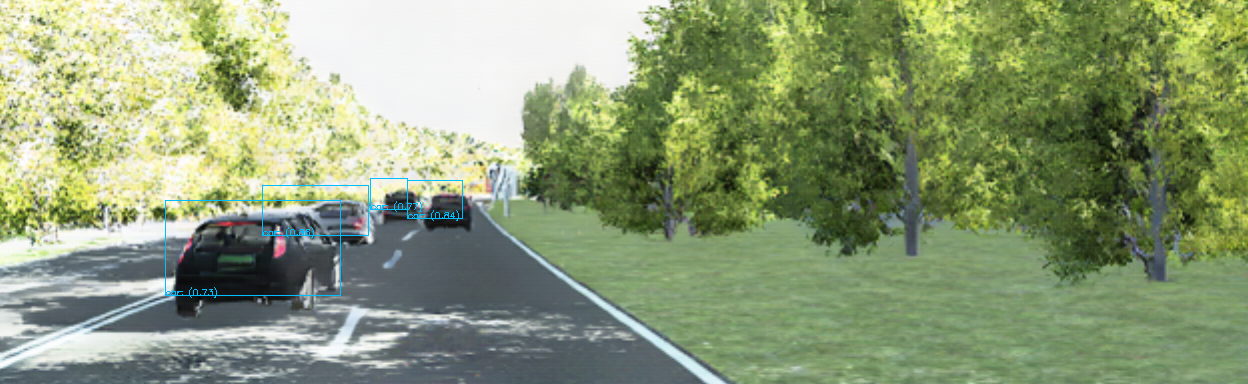}
\end{minipage}
\begin{minipage}[b]{0.5\textwidth}
\includegraphics[width=\textwidth]{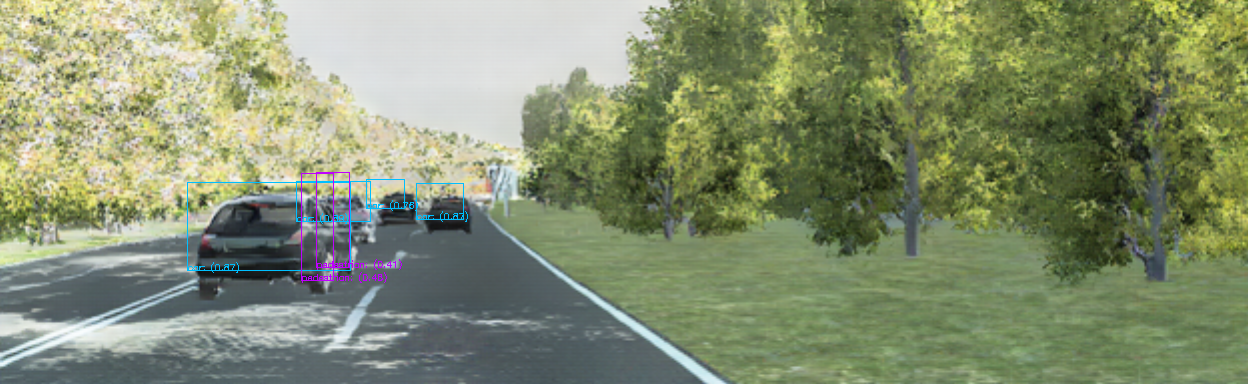}
\end{minipage}
\begin{minipage}[b]{0.5\textwidth}
\vspace{2px}
\includegraphics[width=\textwidth]{figures/benign_semantic_car.png}
\end{minipage}
\begin{minipage}[b]{0.5\textwidth}
\includegraphics[width=\textwidth]{figures/adv_semantic_car.png}
\end{minipage}
\caption{\small \textbf{Semantic space adversarial examples.} Benign re-rendered {VKITTI} image (left), adversarial examples generated by iterative sFGSM over a combination of semantic features (right). Cyan boxes indicate car detected, purple indicated pedestrian, and yellow indicate cyclist. The adversarial example introduces small changes in car positions and orientations, and noticeable changes in their color. This causes the network to detect pedestrians where there are none (top) and to fail to detect a car in the immediate foreground (bottom).}
\end{figure}
\begin{figure}[htpb]
\centering

\includegraphics[scale=0.5]{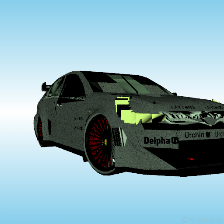}\,%
\includegraphics[scale=0.5]{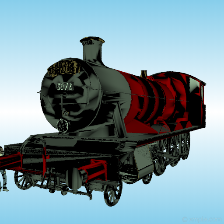}
\includegraphics[scale=0.5]{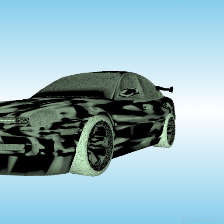}

\includegraphics[scale=0.5]{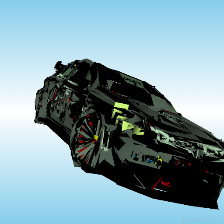}\,%
\includegraphics[scale=0.5]{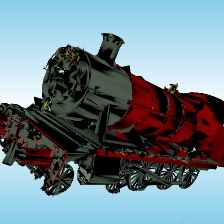}
\includegraphics[scale=0.5]{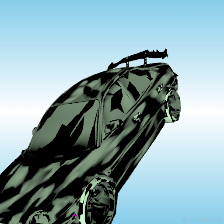}

\caption{\small{Benign images vs their adversarial counterparts generated by attacking the object's pose and vertices and feeding it back into \texttt{redner}. The benign images are on the top row, with the labels "car", "train", and "car", respectively. The adversarial images are on the bottom and are misclassified as "boat", "car", and "helmet".}}
\end{figure}

\end{document}